\documentclass[journal]{IEEEtran}
\usepackage{graphicx}
\usepackage{cite}
\usepackage{picinpar}
\usepackage{amsmath}
\usepackage{url}
\usepackage{flushend}
\usepackage{amsfonts}
\usepackage{colortbl}
\usepackage{soul}
\usepackage{multirow}
\usepackage{pifont}
\usepackage{color}
\usepackage{alltt}
\usepackage[hidelinks]{hyperref}
\usepackage{enumerate}
\usepackage{siunitx}
\usepackage{breakurl}
\usepackage{epstopdf}
\usepackage{pbox}
\usepackage[table]{xcolor}
\usepackage{placeins}
\usepackage{dblfloatfix}
\usepackage{bm}
\usepackage{booktabs}
\usepackage{multirow}
\usepackage{tabu}
\usepackage{dblfloatfix}
\usepackage{array}

\ifCLASSINFOpdf
\else
\fi
\ifCLASSOPTIONcompsoc
  \usepackage[caption=false,font=normalsize,labelfont=sf,textfont=sf]{subfig}
\else
  \usepackage[caption=false,font=footnotesize]{subfig}
\fi
\begin{document}
\title{Collective Decision of One-vs-Rest Networks for Open Set Recognition}

\author{Jaeyeon~Jang and
        Chang Ouk Kim
        % <-this % stops a space
\thanks{This work was supported by the National Research Foundation of Korea (NRF) grant funded by the Korean government (MSIT) (NRF-2019R1A2B5B01070358).}% <-this % stops a space
\thanks{The authors are with the Department of Industrial Engineering, Yonsei University, Seoul 03722, Republic of Korea (e-mail: kimco@yonsei.ac.kr).}
}


\markboth{Journal of \LaTeX\ Class Files,~Vol.~14, No.~8, August~2015}%
{Shell \MakeLowercase{\textit{et al.}}: Bare Demo of IEEEtran.cls for IEEE Journals}



\maketitle

\begin{abstract}
Unknown examples that are unseen during training often appear in real-world machine learning tasks, and an intelligent self-learning system should be able to distinguish between known and unknown examples. Accordingly, open set recognition (OSR), which addresses the problem of classifying knowns and identifying unknowns, has recently been highlighted. However, conventional deep neural networks using a softmax layer are vulnerable to overgeneralization, producing high confidence scores for unknowns. In this paper, we propose a simple OSR method based on the intuition that OSR performance can be maximized by setting strict and sophisticated decision boundaries that reject unknowns while maintaining satisfactory classification performance on knowns. For this purpose, a novel network structure is proposed, in which multiple one-vs-rest networks (OVRNs) follow a convolutional neural network feature extractor. Here, the OVRN is a simple feed-forward neural network that enhances the ability to reject nonmatches by learning class-specific discriminative features. Furthermore, the collective decision score is modeled by combining the multiple decisions reached by the OVRNs to alleviate overgeneralization. Extensive experiments were conducted on various datasets, and the experimental results showed that the proposed method performed significantly better than the state-of-the-art methods by effectively reducing overgeneralization.
\end{abstract}


% \fi
\IEEEpeerreviewmaketitle

\section{Introduction}
\IEEEPARstart{R}{ecent} advancements in deep learning have greatly improved the performance of recognition systems \cite{He2016, Parkhi2015, Simonyan2015, Krizhevsky2012}, which can now surpass human-level performance in terms of classification error rates \cite{He2015}. However, the vast majority of recognition systems are designed under closed-world assumptions, in which all categories are known a priori. Although this assumption holds in many applications, the need to detect objects unknown at training time while classifying knowns, which is called \textit{open set recognition} (OSR), has recently been highlighted \cite{S, Jung2020, Henrydoss2017, Cruz2017}. Furthermore, the ability to distinguish between known and unknown has been considered a key element of intelligent self-learning systems \cite{Boult2019}.

\begin{figure}[h]\centering
  \includegraphics[width=8.5cm]{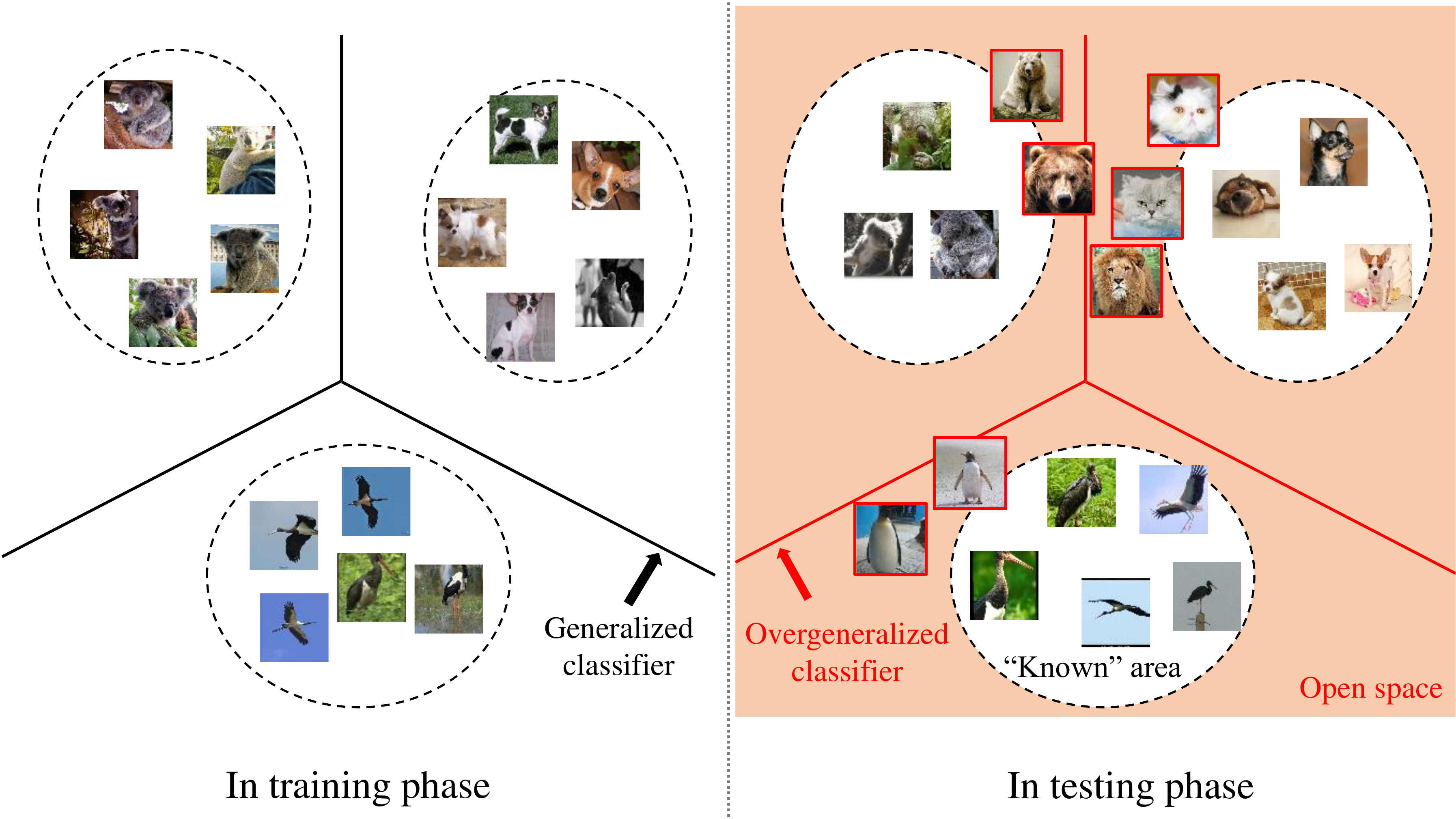}
  \caption{Open set scenario where only “koala”, “dog”, and “black stork” examples are provided during training, but “bear”, “cat”, “lion”, and “penguin”, unseen during training, appear during testing.}
  \label{Open set scenario}
%	\vspace{-10pt}
\end{figure}

Fig. \ref{Open set scenario} illustrates a typical example of an open set scenario. Most recognition systems assume a closed set scenario, in which only samples of known classes appear during the testing stage. If this assumption holds, setting generalized decision boundaries is effective because they can reduce model overfitting, simultaneously reducing the possibility of misclassification \cite{Moosavi-Dezfooli2019, Spigler2019}. However, in open set scenarios, a recognition system designed assuming a closed set results in \textit{overgeneralization}, creating a large \textit{open space}. Here, open space refers to the positively labeled space sufficiently far from any known training samples. Thus, in addition to high closed set classification performance, the purpose of OSR is to minimize overgeneralization to reject samples in open space.

Many deep learning-based OSR methods have recently been proposed with the goal of mitigating overgeneralization \cite{Spigler2019, Bendale2016, Shu2017, Yoshihashi2019, Oza2019a, Oza2019, Sun2020, Ge2017, Neal2018}. Most OSR methods aim to provide regularized probability scores that can reject unknowns while classifying known samples. For this purpose, post score analysis \cite{Scheirer2011} has been applied to the recognition scores or latent features produced by learned networks. Furthermore, recently, some researchers have applied autoencoders \cite{Spigler2019, Yoshihashi2019, Oza2019a, Oza2019, Sun2020} and generative adversarial networks (GANs) \cite{Ge2017, Neal2018} to provide more discriminative information for unknown identification, in addition to deep neural networks (DNNs) using softmax layers. However, these reconstructive and generative networks can represent only a limited portion of unknowns because the diversity of unknowns is infinite. Thus, the key for OSR is to tighten the decision boundaries for known classes while maintaining satisfactory classification performance on known classes. However, the commonly used softmax activation is very weak from this perspective because softmax is designed to measure the relative likelihood of a known class compared to other known classes, yielding overgeneralized decision boundaries only for known classes.

In this paper, we propose a new network structure in which a set of one-vs-rest networks (OVRNs) follows a convolutional neural network (CNN) feature extractor. Here, an OVRN is a simple feed-forward neural network using sigmoid output activation. Each OVRN learns class-specific discriminative features that distinguish between matches and nonmatches individually for each class, thereby enhancing the ability to reject nonmatches. Accordingly, the proposed network applying OVRNs can assign lower confidence scores to unknown samples than softmax CNN classifiers can. Here, the confidence score is the likelihood of belonging to a known class. In addition, given samples, the OVRNs of nontarget classes can provide useful information. For example, the lower the output of a nontarget class OVRN is, the more likely it is the sample belongs to the target class. Thus, we propose a collective decision method that combines the multiple decisions reached by OVRNs to establish more sophisticated decision boundaries that reduce the open space.

Comprehensive experiments were conducted on various datasets. The experimental results showed that the proposed method effectively mitigates overgeneralization. As a result, the proposed method outperformed state-of-the-art OSR methods in various open set scenarios, despite the simplicity of the implementation.

The remainder of this paper is organized as follows. Section \ref{Chap2} presents related work on existing OSR methods. In Section \ref{Chap3}, we provide details of the proposed method. Section \ref{Chap4} verifies the proposed method from various perspectives. Finally, in Section \ref{Chap5}, we conclude this study and present future research plans.

\section{Related Works} \label{Chap2}
OSR has been systematically studied since the early 2010s. In the early days of its history, shallow machine learning model-based approaches were at the forefront of advancement \cite{Scheirer2013, Cevikalp2017, Scheirer2014, Jain2014, Zhang2017, Bendale2015, MendesJunior2017, Rudd2018}. During this period, many researchers redesigned traditional machine learning models, including support vector machines, to minimize open set risk \cite{Scheirer2013}, which comprises open space risk and empirical risk. Here, open space risk represents the risk of misclassifying an unknown example in open space as belonging to a known class, while empirical risk represents the loss of classification performance for known data.

With the advancement of deep learning techniques, many deep learning-based OSR methods have been proposed. Most deep learning-based approaches aim to alleviate the overgeneralization effect of DNNs, which generally use a softmax output layer \cite{Bendale2016, Shu2017, Yoshihashi2019, Ge2017, Neal2018}. The first deep model introduced for OSR was OpenMax \cite{Bendale2016}, which models the class-specific distribution of activation vectors in the penultimate layer of a CNN and transforms the distributional information into decision scores. Shu \textit{et al.} \cite{Shu2017} proposed a deep open classification (DOC) network, which is a CNN with a sigmoid output layer that adopts the concept of multitask learning \cite{Xu2019}. They showed that the DOC can further reduce open space risk by tightening the decision boundaries via Gaussian fitting. Yoshihashi \textit{et al.} \cite{Yoshihashi2019} proposed a classification-reconstruction learning algorithm for OSR (CROSR) that implements both classification and reconstruction simultaneously. They additionally utilized hierarchical latent representation in the OpenMax score calculation, showing that robust unknown detection is possible.

Some researchers have applied adversarial learning to generate samples to account for open space \cite{Ge2017, Neal2018}. Ge \textit{et al.} \cite{Ge2017} proposed a generative OpenMax (G-OpenMax) model that learns synthetic unknown samples generated by a conditional GAN (CGAN) to enhance the rejection capability. Here, CGAN generated samples based on a mixture of class labels. Neal \textit{et al.} \cite{Neal2018} proposed an OSR method using counterfactual image (OSRCI), where an encoder-decoder GAN was used to produce a counterfactual sample that is close to a known sample in the latent space but does not belong to any of the known classes. However, the two GAN-based methods produce synthetic samples limited to only a small subspace of unknowns.

Recently, to enhance the ability to detect unknowns, researchers have suggested two-stage methods that detect unknowns first and then classify samples identified as known using a typical softmax CNN \cite{Oza2019a, Oza2019, Sun2020}. They assumed that an open set system maximizes performance by identifying unknown samples well because high-performance classification on the samples detected as known can easily be achieved by adopting a state-of-the-art closed-set classifier. However, the classification performance of certain classes with high intraclass diversity can be significantly reduced because this class is likely to have a wide positive area for unknowns when the unknown samples are misclassified by the unknown detection model.

\section{Proposed Method} \label{Chap3}
In this section, we propose a collective decision method based on OVRNs. Here, each OVRN uses sigmoid output activation to learn more discriminative features than the learned using the general softmax output layer. Thus, we first address the difference between using softmax and sigmoid activation for the output layer in an open set scenario. Then, a detailed description of OVRN is provided, along with the structural advantages. Finally, we describe the recognition rule based on collective decisions obtained by combining the decisions reached by multiple OVRNs.

\subsection{Sigmoid vs. Softmax}
Most DNN classifiers use a softmax output layer to learn categorical distributions. When $M$ known classes $\bm{y}\in \mathbb{R}^M$ are given, a DNN that uses a softmax output layer learns the probability distribution over $M$ known classes for an input $\bm{x}$. For the $i$-th class $y_i$, the softmax layer produces the following conditional probability:
\begin{gather} 
P(y_i\mid l_{y_1},…,l_{y_M})=\frac{exp(l_{y_i})}{\sum_{m=1}^M exp(l_{y_m})}, \label{eq1}
\end{gather}
where $l_{y_i}$ is the logit of class $y_i$.

\begin{figure}[h]\centering
  \includegraphics[width=8.5cm]{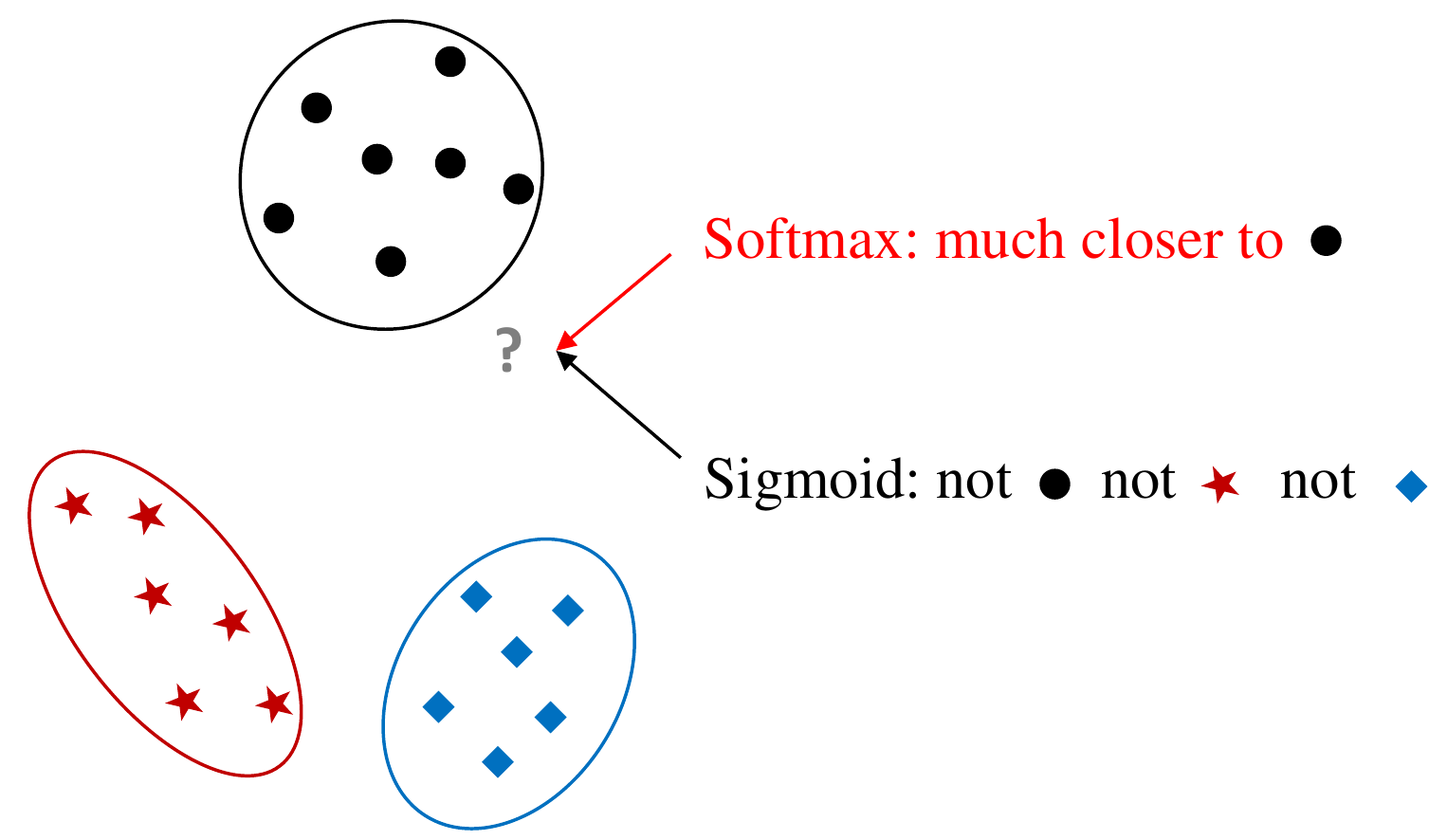}
  \caption{Difference between softmax and sigmoid activation.}
  \label{Sigmoid}
%	\vspace{-10pt}
\end{figure}

A network using a softmax output layer is trained to increase $l_{y_i}$ relative to the logits of the other classes. This training mechanism causes the activations of nontarget classes to converge to zero by increasing the logit of the target class in the normalization term (the denominator), even though the logits of the nontarget classes do not decrease substantially \cite{Boult2019, Oland2017}. The activations that converge to zero backpropagate very small gradients through the network, and the network rarely learns latent representations that can be useful for discriminating nonmatch samples. In particular, since the softmax function is designed to measure the relative likelihood of a known class compared to the other known classes, the softmax layer gives a high confidence score to unknown samples by identifying the most similar class among all known classes. For example, in Fig. \ref{Sigmoid}, softmax is highly likely to determine unknown “?” as a black circle class because the example is much closer to the black circle class than it is to the other classes.

\begin{figure}[h]
\centering
\subfloat[CNN-SoftMax]{\includegraphics[width=0.8\linewidth]{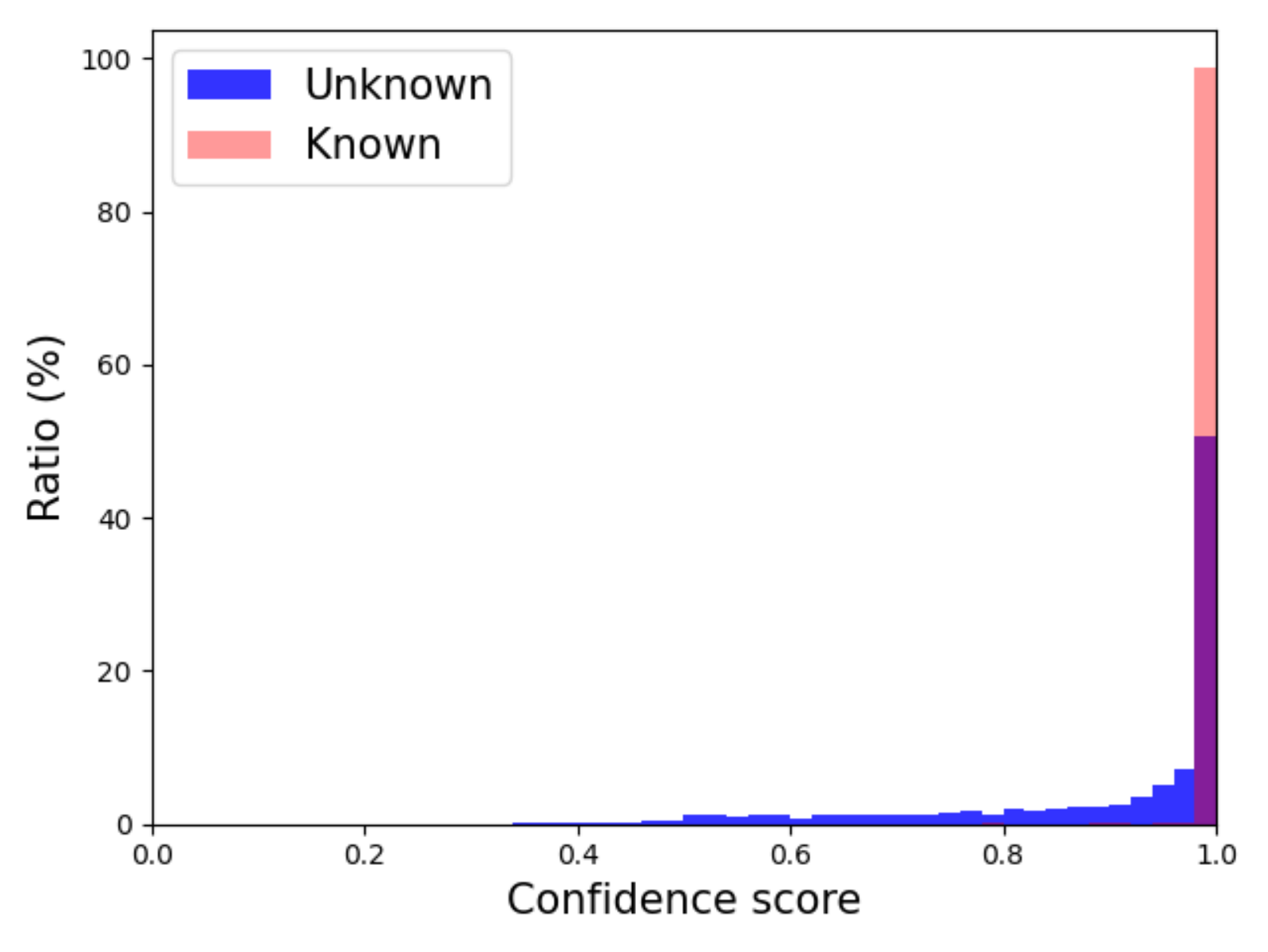}}
\par
\subfloat[CNN-Sigmoid]{\includegraphics[width=0.8\linewidth]{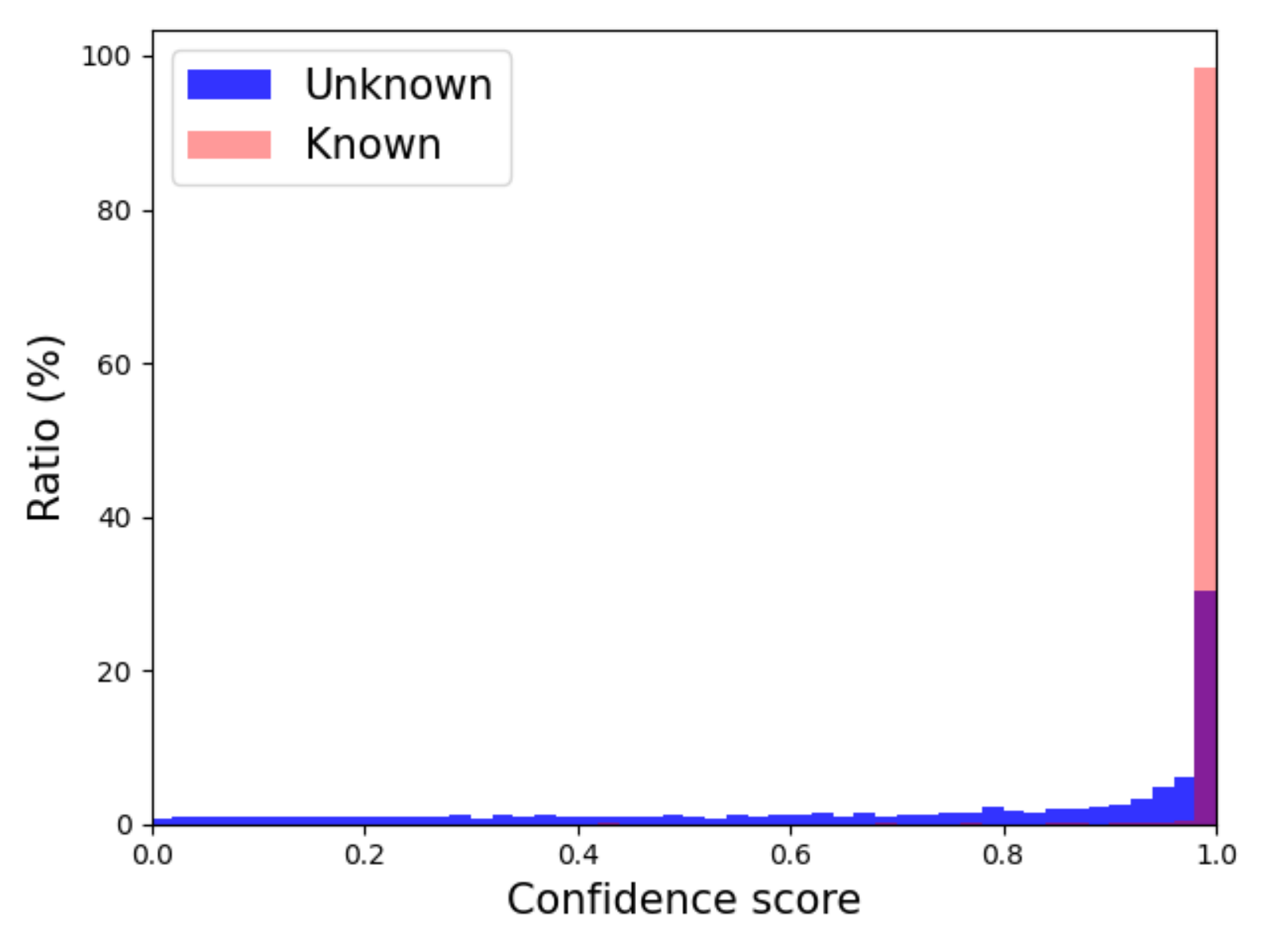}}
\caption{Distribution of confidence scores for known and unknown classes on MNIST. The maximum posterior probability among known classes was used as the confidence score.}
\label{Comparison1}
\end{figure}

On the other hand, if sigmoid activation is applied to the output layer, the sigmoid layer yields the probability for $y_i$, which is conditioned only on $l_{y_i}$, as follows:
\begin{gather} 
P(y_i\mid l_{y_i})=\frac{1}{1+exp(-l_{y_i})}. \label{eq2}
\end{gather}
Thus, each sigmoid output node is individually trained. Additionally, in contrast to softmax activation, match and nonmatch examples are learned equally through each sigmoid output node during training. Thus, each sigmoid node is trained to discriminate a dissimilar example from match examples. As a result, networks using the sigmoid output layer provide lower confidence scores for unknowns than do conventional networks using the softmax output layer, as shown in Fig. \ref{Comparison1}. In the figure, MNIST \cite{Web2012} was partitioned into six known classes $(0\sim 5)$ and four unknown classes $(6\sim 9)$.

When sigmoid activation is applied, each output node provides individual determination. For example, in Fig. \ref{Sigmoid}, the sigmoid layer can provide three individual determinations. In addition, in the presence of unknown samples, the posterior probability of each class becomes supporting information. For instance, high posterior probabilities for multiple classes suggest that the example belongs to an unknown class.

\subsection{One-vs-Rest Networks}
\begin{figure*}[h]\centering
  \includegraphics[width=11cm]{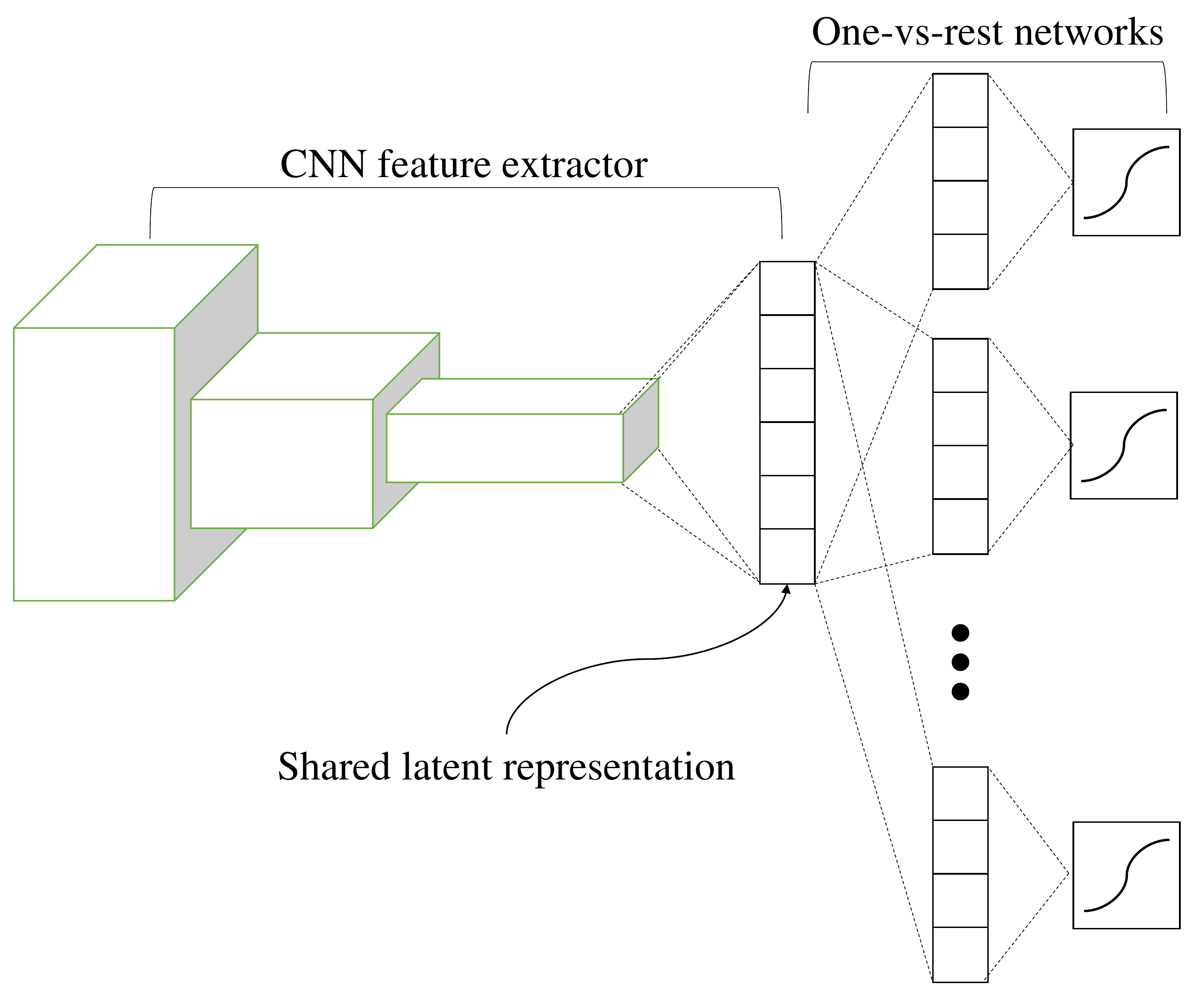}
  \caption{Structure of the proposed network.}
  \label{OVRN}
\end{figure*}

For the proposed method, the general softmax output layer is replaced with a set of OVRNs. Accordingly, the proposed network is composed of a CNN feature extractor and the following OVRNs, as shown in Fig. \ref{OVRN}. Let $\mathcal{F}$ be the CNN feature extractor and $\mathcal{G}_i$ be the OVRN for class $y_i$. Then, the posterior probability of $\bm{x}_j$ belonging to $y_i$ is defined as shown in \eqref{eq3}. For each target class, one network with hidden layers and sigmoid output is constructed on the shared latent representation. Thus, compared to a single sigmoid layer, more specialized discriminative information can be learned after the shared latent representation for each target class.
\begin{gather} 
P(y_i\mid\bm{x}_j)=\mathcal{G}_i(\mathcal{F}(\bm{x}_j)). \label{eq3}
\end{gather}

To produce $M$ individual decisions for $M$ known classes, the proposed network is trained to minimize the following binary cross-entropy:
\begin{equation}\label{eq4}
\begin{aligned}
\mathcal{L}(\bm{\theta}_{\mathcal{F}},\bm{\theta}_{\mathcal{G}_1}, ..., &\bm{\theta}_{\mathcal{G}_M})=-\frac{1}{N}\sum_{j=1}^N\sum_{i=1}^M\mathbb{I}(t_j=y_i)\\P(y_i\mid\bm{x}_j)
&+\mathbb{I}(t_j\neq y_i)(1-P(y_i\mid\bm{x}_j)),
\end{aligned}
\end{equation}
where $N$ is the batch size, $\mathbb{I}$ is the indicator function, $t_j$ is the ground-truth label of sample $\bm{x}_j$, and $\bm{\theta}_{\mathcal{F}},\bm{\theta}_{\mathcal{G}_1}, ..., \bm{\theta}_{\mathcal{G}_M}$ represent the parameters of network $\mathcal{F},\mathcal{G}_1,…,\mathcal{G}_M$.

\subsection{Recognition Rule Based on Collective Decisions}
In this section, we propose a recognition rule based on the collective decisions of OVRNs. For a sample of class $y_i$, the sample is more likely to belong to the target class when the sample has high posterior probability for the target class and low posterior probabilities for all other classes. However, because the sigmoid function scales significantly when the input is relatively low or high, most nontarget OVRNs usually produce zero probability for a sample. Thus, the collective decision score is computed based on the logit of the sigmoid activation. Let $l_{jy_i}$ be the logit value of a sample $\bm{x}_j$ for class $y_i$. Then, $cds_{jy_i}$, the collective decision score for class $y_i$, is computed by the following simple function:
\begin{gather} 
cds_{jy_i}=l_{jy_i}-\frac{1}{M-1}\sum_{m\neq i}l_{jy_m}. \label{eq5}
\end{gather}

Finally, we propose a recognition rule using collective decision scores for both closed set classification and unknown detection, as follows:
\newcommand{\argmax}{\arg\!\max}
\begin{gather} 
y^*= \begin{cases}
    \argmax_{y_i\in \{y_1, ..., y_M \}}cds_{jy_i} & \text{if $cds_{jy_i} > \epsilon_{y_i}$}\\
    \text{"unknown"} & \text{otherwise}
  \end{cases}, \label{eq6}
\end{gather}
where $\epsilon_{y_i}$ is the collective decision score threshold for class $y_i$. $\epsilon_{y_i}$ can easily be set based on collective decision scores of the training data by class. In this paper, we obtain $\epsilon_{y_i}$ to ensure that 95\% of the class $y_i$ training data are recognized as known and classified as the target class.

\section{Experiments}\label{Chap4}

In this section, we conduct extensive experiments, including ablation studies, sensitivity analysis, and comparisons with state-of-the art OSR methods, on seven datasets: MNIST \cite{Web2012}, EMNIST \cite{Cohen2017}, Omniglot \cite{Lake2015}, CIFAR-10 \cite{Krizhevsky2009a}, CIFAR-100 \cite{Krizhevsky2009a}, ImageNet \cite{Russakovsky2015}, and LSUN \cite{Yu2015}. To measure the performance in an open set scenario, the macroaveraged F1 score ($F_1$) in the known classes and “unknown” was used. In the OSR problem, the ratio of unknown classes to known classes affects the classification performance. Thus, openness was introduced to measure how open the problem setting is \cite{Scheirer2013}:
\begin{gather} 
\text{openness}=1-\sqrt{\frac{2C_T}{C_E+C_R}}, \label{eq7}
\end{gather}
where $C_T$ is the number of classes used in training, $C_E$ is the number of classes used in evaluation (testing), and $C_R$ is the number of classes to be recognized.

\begin{figure*}[b]
\centering
\subfloat[CNN-SoftMax]{\includegraphics[width=0.33\linewidth]{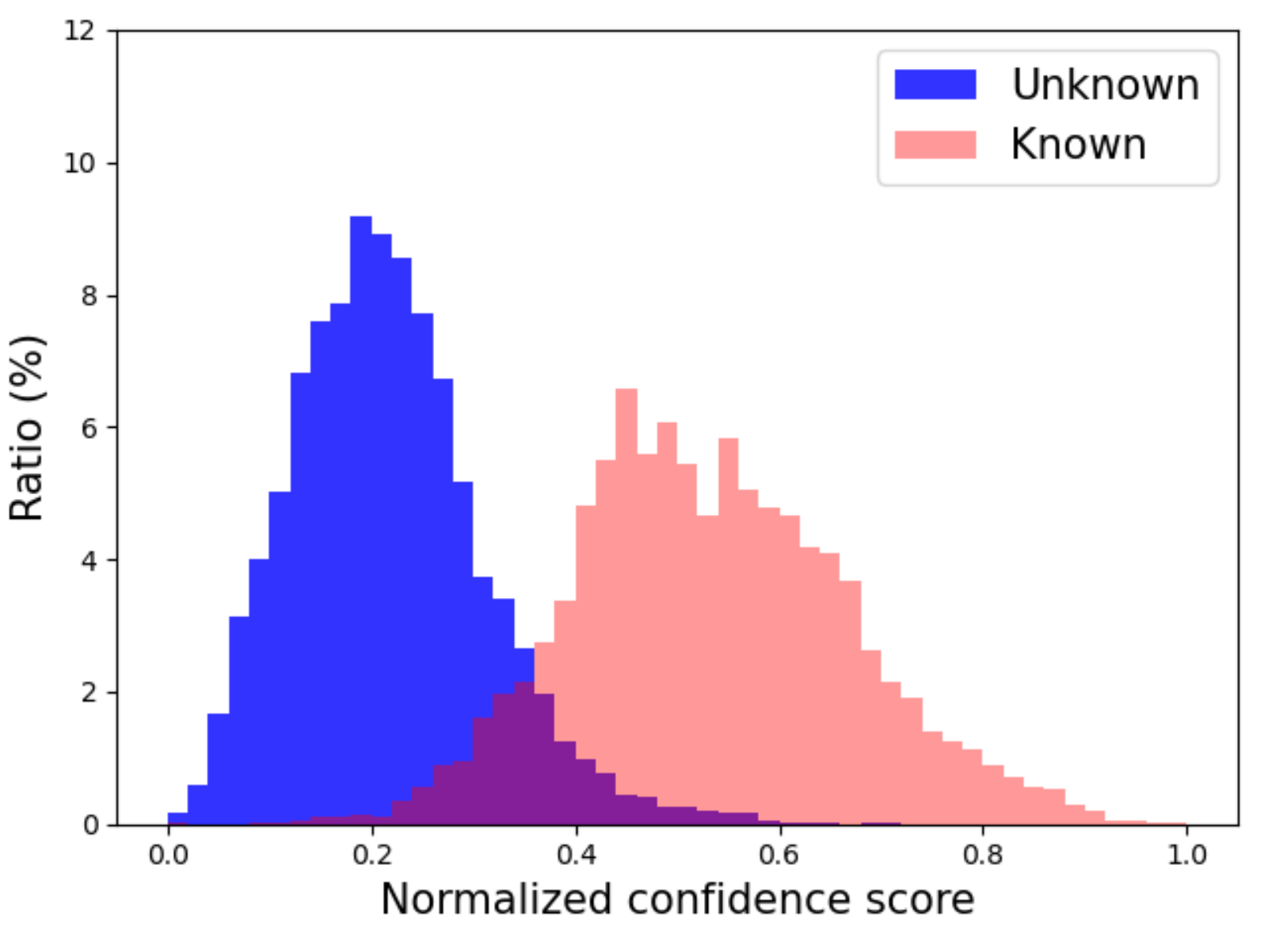}}
\subfloat[CNN-Sigmoid]{\includegraphics[width=0.33\linewidth]{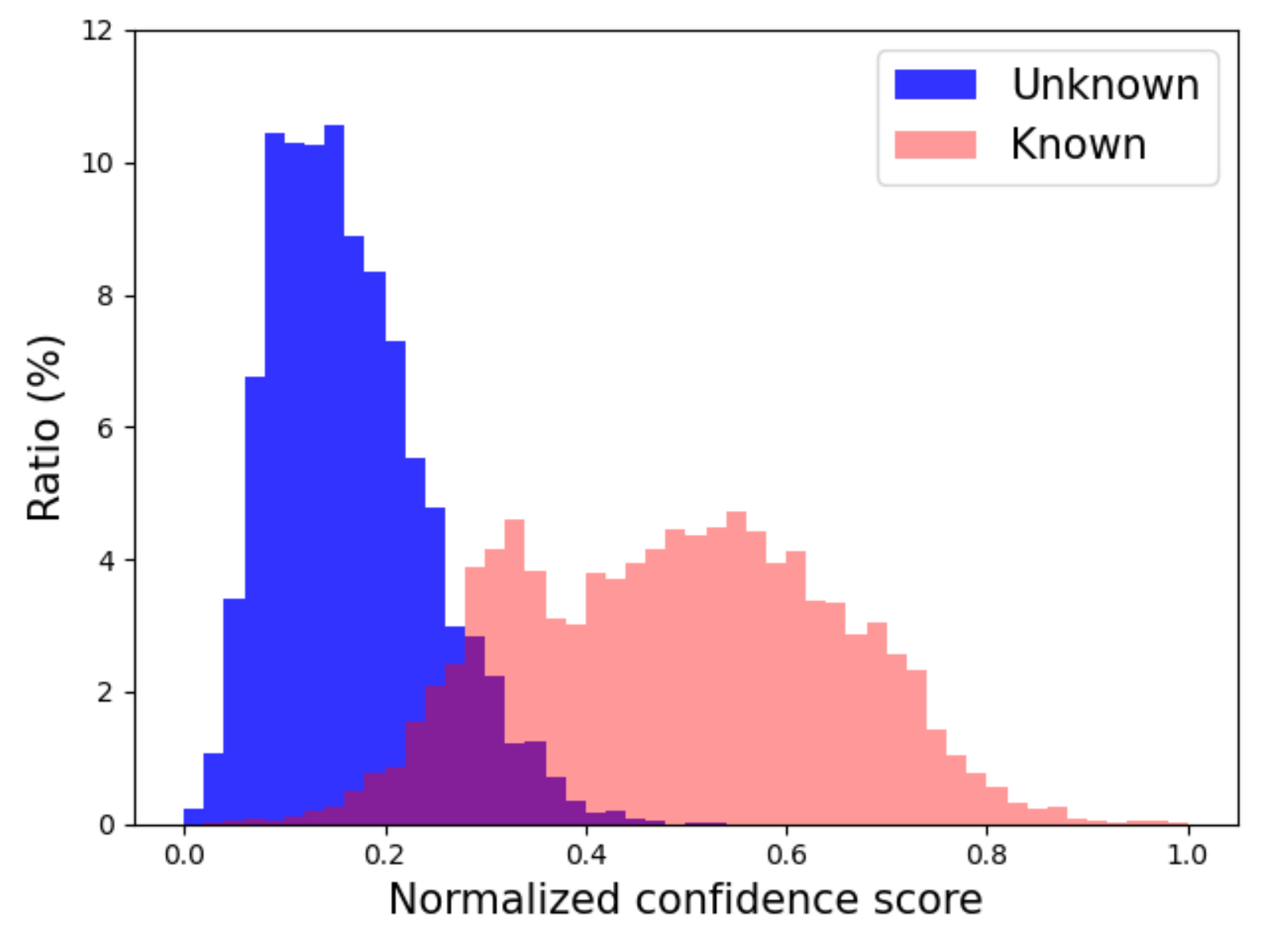}}
\subfloat[CNN-OVRN]{\includegraphics[width=0.33\linewidth]{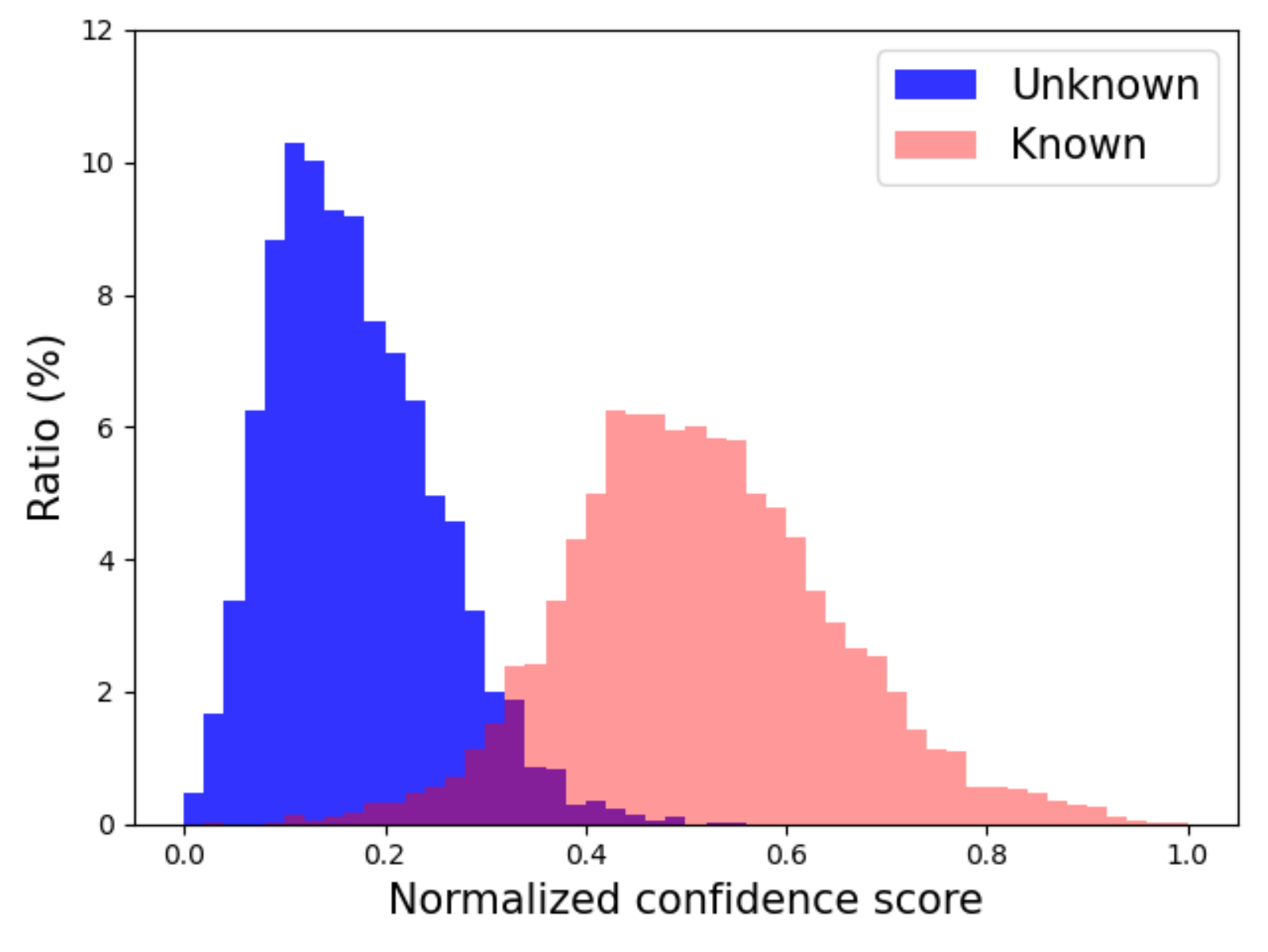}}
\caption{Distribution of collective decision scores for known and unknown classes on MNIST. The maximum score among known classes was normalized on a zero-one scale and used as the confidence score.}
\label{Comparison2}
\end{figure*}

\begin{figure*}[b]
\centering
\subfloat[Known: MNIST, Unknown: EMNIST]{\includegraphics[width=0.4\linewidth]{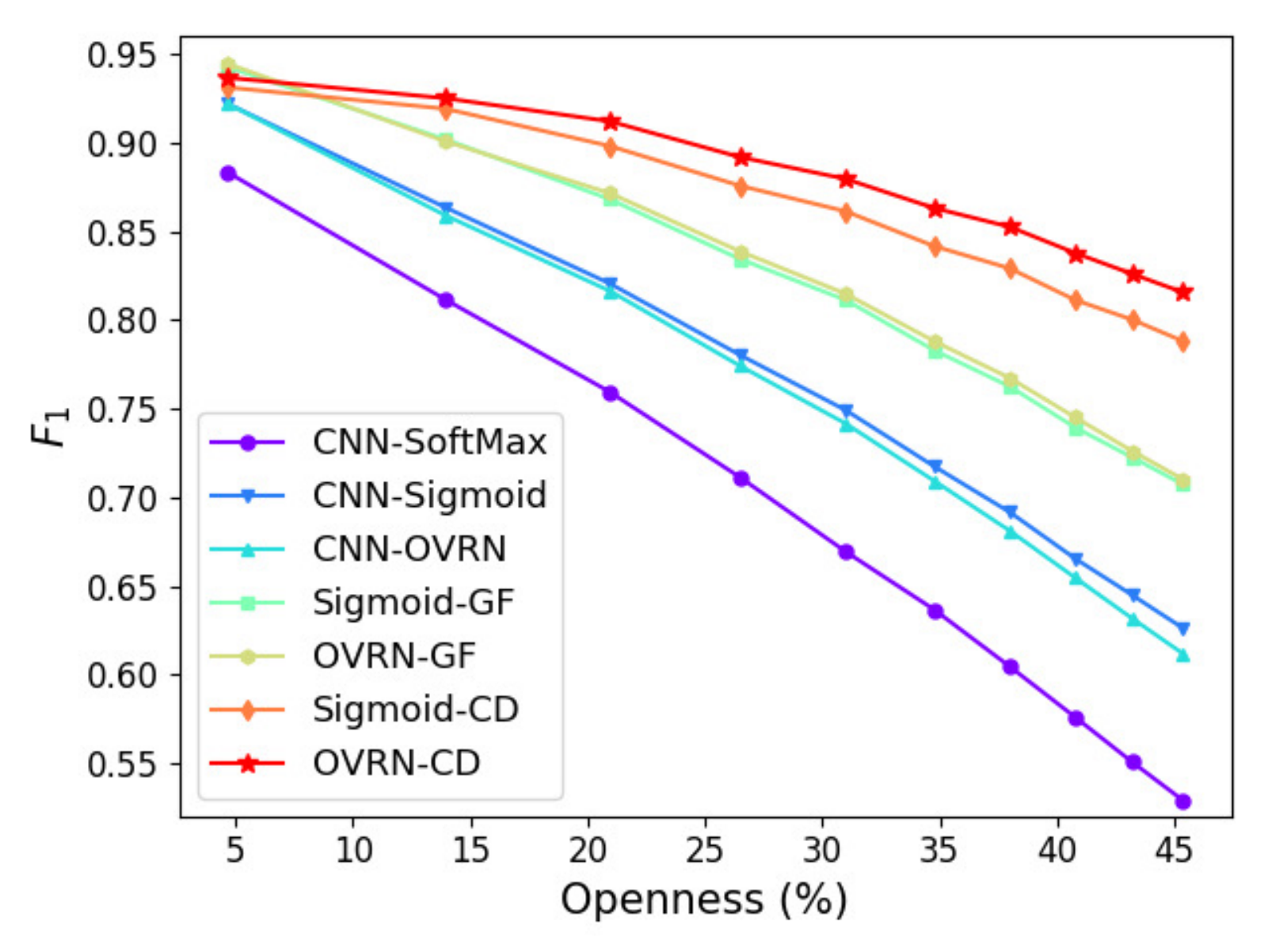}}
\subfloat[Known: CIFAR-10, Unknown: CIFAR-100]{\includegraphics[width=0.4\linewidth]{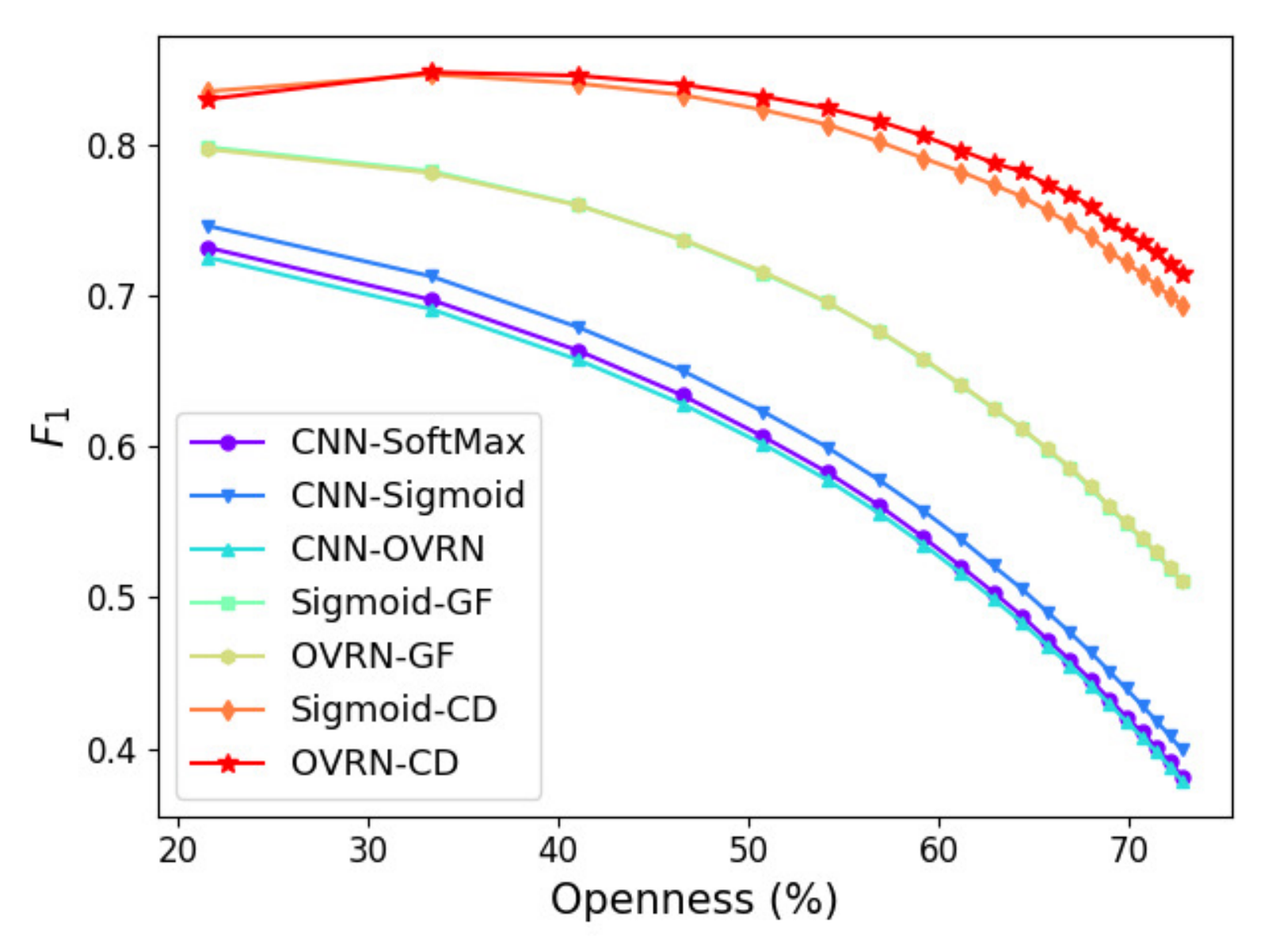}}
\caption{F1 scores against varying openness for seven methods.}
\label{Ablation}
\end{figure*}

For backbone networks (CNN feature extractor in this paper), plain CNN and the redesigned VGGNet defined in \cite{Yoshihashi2019} were employed. Specifically, the plain CNN combined with single hidden layer OVRNs consisting of 64 hidden nodes was used to train on MNIST. When other datasets were used for training, the redesigned VGGNet combined with single hidden layer OVRNs consisting of 128 hidden nodes was used. ReLU activation was applied to the hidden layers. We used the Adam optimizer with a learning rate of 0.002 to train all the networks used in the experiments.

\subsection{Ablation Study}
First, we conducted qualitative analysis to validate the effectiveness of applying OVRN when the collective decision method is applied. For the analysis, the MNIST dataset was partitioned into six known classes $(0\sim 5)$ and four unknown classes $(6\sim 9)$. Fig. \ref{Comparison2} shows histograms of the normalized confidence scores for the known and unknown classes. By extracting collective decision scores, unknown samples were separated more clearly from known samples compared to the results in Fig. \ref{Comparison1}. In addition, CNN-OVRN provided the best-separated histogram, revealing that unknowns can be rejected more easily than when using CNN-SoftMax and CNN-Sigmoid.

\begin{figure*}[t]
\centering
\subfloat[Known: MNIST, Unknown: EMNIST]{\includegraphics[width=0.4\linewidth]{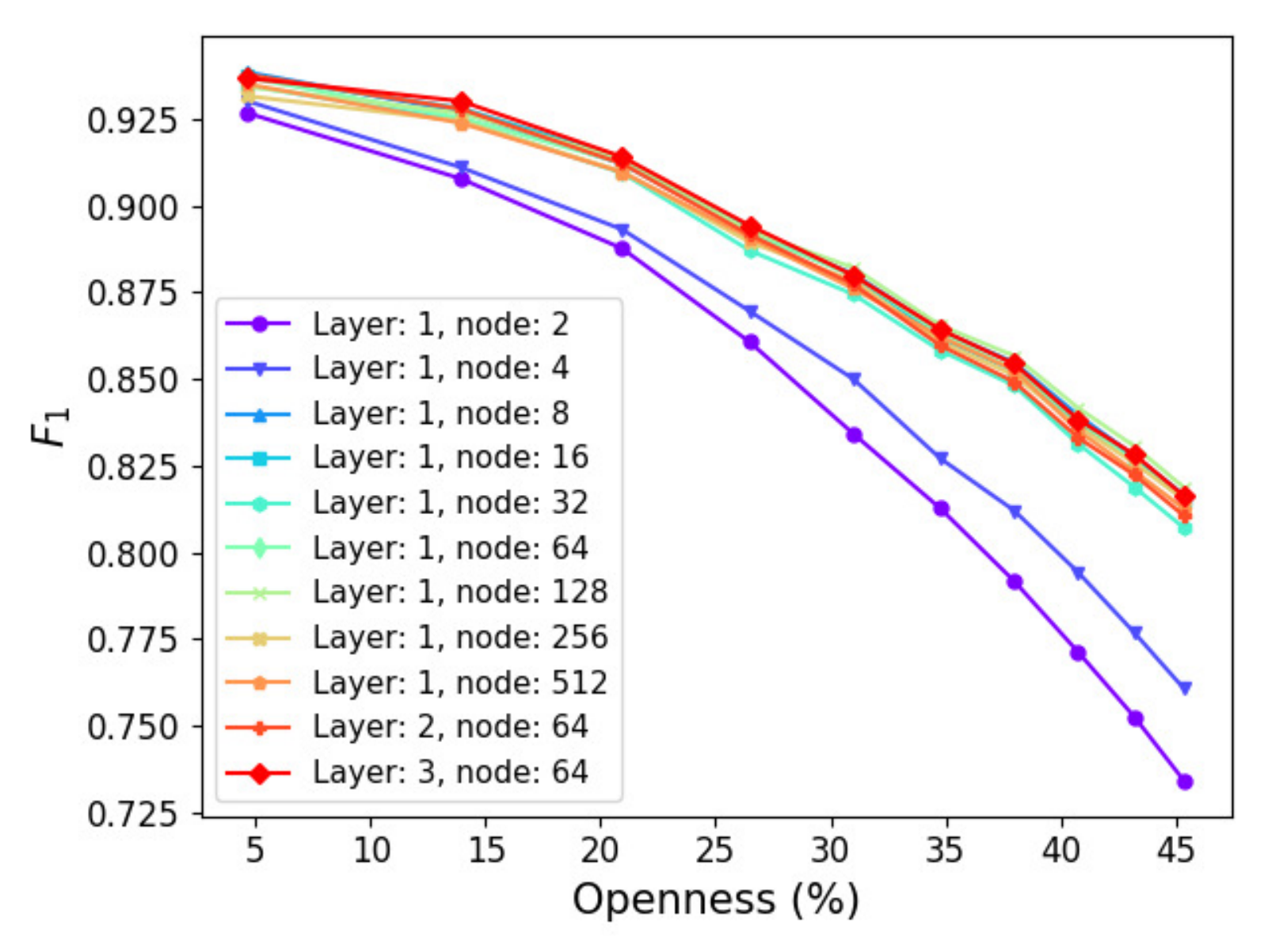}}
\subfloat[Known: CIFAR-10, Unknown: CIFAR-100]{\includegraphics[width=0.4\linewidth]{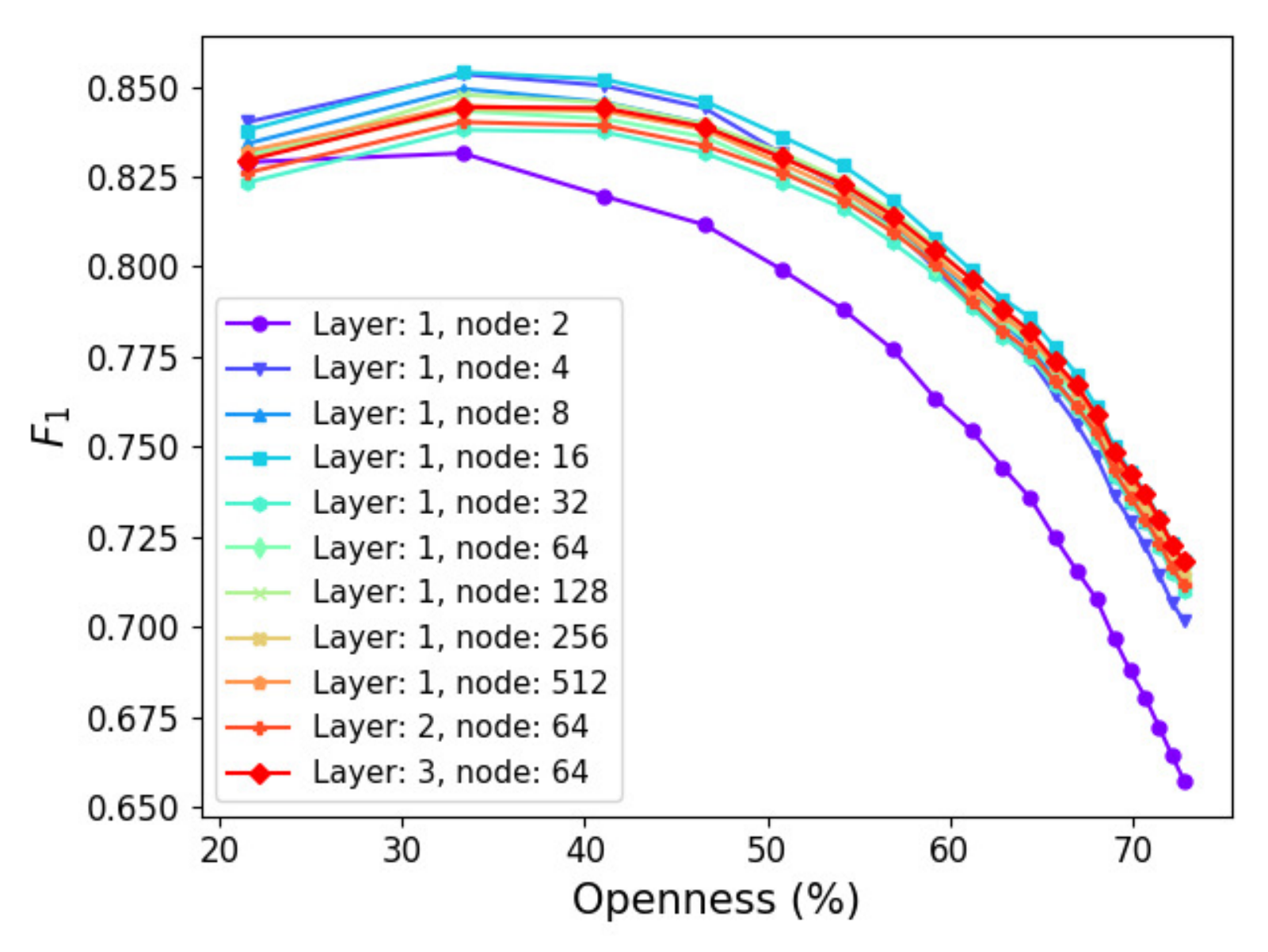}}
\caption{F1 scores according to the number of hidden layers and nodes in OVRNs.}
\label{Sensitivity}
\end{figure*}

To analyze the quantitative contributions of introducing the OVRNs and the collective decision method, we compared the following seven baselines:
\begin{enumerate}[1)]
\item CNN-SoftMax: The first baseline model is a conventional softmax CNN. The threshold for unknown detection is set to 0.5, which is the most commonly used value for the classwise rejection threshold. That is, a sample is rejected as unknown if the sample has an output score of less than 0.5 for all known classes.
\item CNN-Sigmoid: In this baseline, the softmax output layer in CNN-SoftMax is replaced by a sigmoid layer. Unknown detection is implemented just as it is in CNN-SoftMax.
\item CNN-OVRN: OVRNs are applied instead of the single sigmoid output layer. The training and testing procedures are the same as those in CNN-Sigmoid.
\item Sigmoid-GF: A Gaussian fitting (GF)-based threshold setting is added to baseline 2. This model is the DOC, which was proposed for OSR applications in the field of natural language processing \cite{Shu2017}.
\item OVRN-GF: A GF-based threshold setting is added to baseline 3.
\item Sigmoid-CD: A collective decision-based recognition rule is incorporated into CNN-Sigmoid.
\item OVRN-CD (proposed method): A collective decision-based recognition rule is incorporated into CNN-OVRN.
\end{enumerate}

For the ablation study, two experimental settings were considered. In the first setting, 10 digit classes of MNIST and 47 letter classes of EMNIST were used as knowns and unknowns, respectively. Openness was varied from 4.7\% to 45.4\% by randomly sampling 2 to 47 unknown classes in intervals of 5. In the second setting, four non-animal classes of CIFAR-10 and hundred classes of CIFAR-100 were used as knowns and unknowns, respectively. We varied openness from 21.6\% to 72.8\% by randomly sampling 5 to 100 unknown classes in intervals of 5. For each experimental setting, the performances of five randomized unknown class samplings were averaged, except in the case where all unknown classes were used.

The experimental results are shown in Fig. \ref{Ablation}. As openness increases, the F1 scores of all baseline models generally decrease. Among the baselines, the proposed OVRN-CD achieved the best performance, except in the lowest openness case, thereby showing the most robust results. CNN-Sigmoid performed better than CNN-OVRN when no calibrations were conducted for the confidence score. However, when the collective decision method was incorporated, OVRN-CD performed better as openness increased. Specifically, for the highest openness case, the collective decision method improved performance by 0.336 for CNN-OVRN when the CIFAR-100 dataset was considered unknown. This result reveals that the proposed network structure can achieve synergy through combination with the collective decision method.

\subsection{Sensitivity Analysis}
We conducted a sensitivity analysis on the number of hidden layers and nodes in the OVRN. For the sensitivity analysis, the two experimental designs used for the ablation study were adopted. Fig. \ref{Sensitivity} shows that if an OVRN has eight or more hidden nodes, the performance improvement is not significant, even if more nodes are added. Furthermore, the addition of hidden layers does not have a significant effect. That is, an OVRN only needs to provide minimal complexity.

\subsection{Comparison with State-of-the-Art Methods}
\begin{figure}[t]\centering
  \includegraphics[width=8cm]{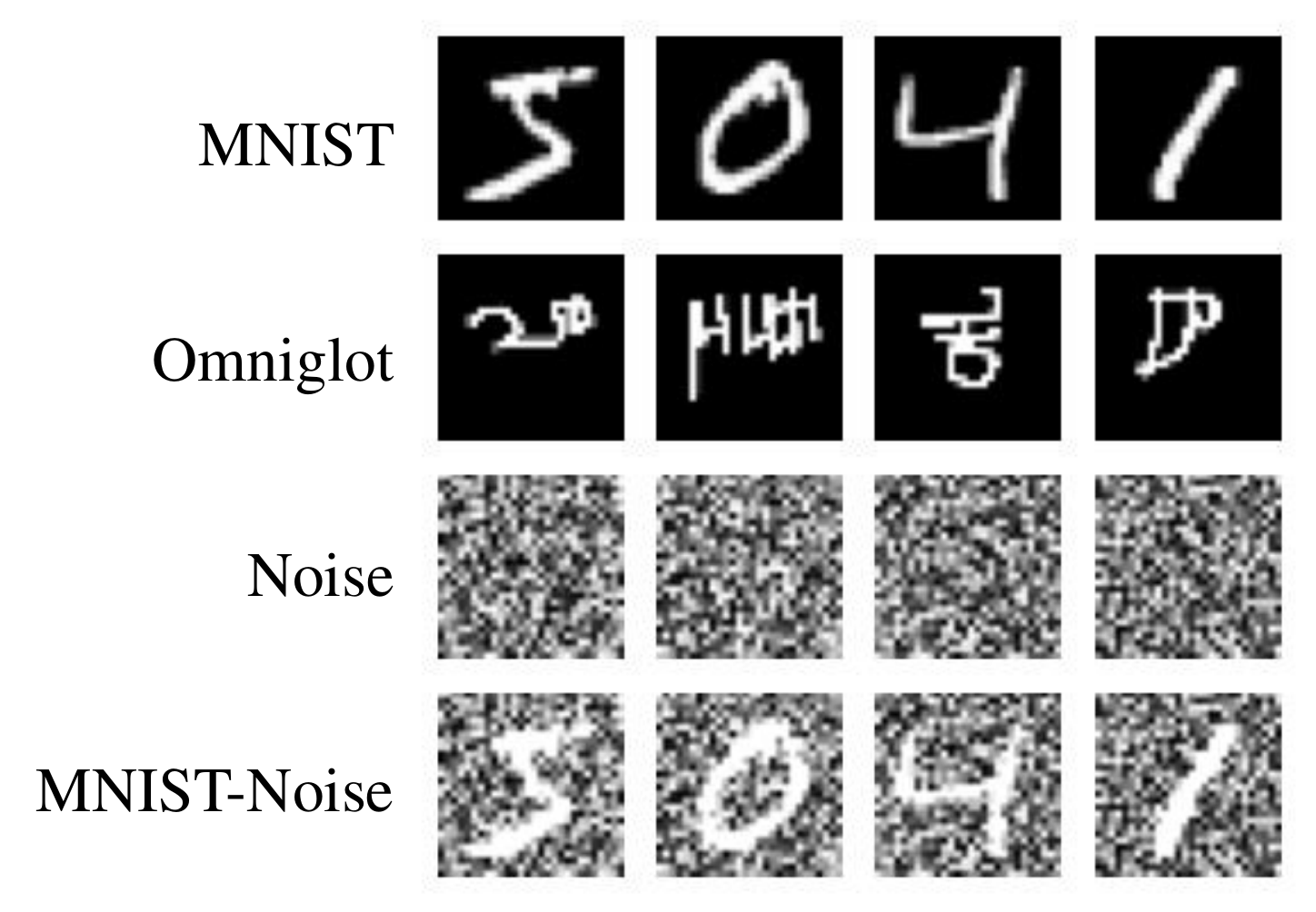}
  \caption{Samples from MNIST, Omniglot, Noise, and MNIST-Noise.}
  \label{outliers}
\end{figure}

Finally, we compared the proposed method with the state-of-the-art methods in terms of OSR performance. The training samples of the MNIST or the CIFAR-10 were used for training, but for testing, another dataset was added to the MNIST or the CIFAR-10 test samples as unknowns. Specifically, when the MNIST was trained, we used three datasets of grayscale images, Omniglot, Noise, and MNIST-Noise, as the unknowns (see Fig. \ref{outliers}). Here, the noise is a set of images synthesized by independently sampling each pixel value from a uniform distribution on [0, 1], and MNIST-Noise is a synthesized dataset created by superimposing the test images of the MNIST on the noise. Each “unknown” dataset contains 10,000 samples, the same as the MNIST test dataset, making the known to unknown ratio 1:1. The comparison results are shown in Table \ref{table_1}. Details about LadderNet and DHRNet can be found in \cite{Valpola2015} and \cite{Yoshihashi2019}. The proposed OVRN-CD outperformed the state-of-the-art methods on all “unknown” datasets, showing that the OSR performance can be dramatically improved simply by introducing a collective decision method based on OVRNs.

\begin{table}[t]
\caption{Open Set Recognition Results on MNIST with Various “Unknown” Datasets}
\label{table_1}
\centering
\begin{tabular}{l|c|c|c}
\hline\hline
Method & Omniglot & MNIST-Noise & Noise\\
\hline									
Softmax & 0.592 & 0.641 & 0.826\\
OpenMax \cite{Bendale2016} & 0.680 & 0.720 & 0.890\\
LadderNet+Softmax \cite{Yoshihashi2019} & 0.588 & 0.772 & 0.828\\
LadderNet+OpenMax \cite{Yoshihashi2019} & 0.764 & 0.821 & 0.826\\
DHRNet+Softmax \cite{Yoshihashi2019} & 0.595 & 0.801 & 0.829\\
DHRNet+OpenMax \cite{Yoshihashi2019} & 0.780 & 0.816 & 0.826\\
CROSR \cite{Yoshihashi2019} & 0.793 & 0.827 & 0.826\\
DOC \cite{Shu2017} & 0.863 & 0.892 & 0.921\\
CGDL \cite{Sun2020} & 0.850 & 0.887 & 0.859\\
OVRN-CD (Ours) & \textbf{0.918} & \textbf{0.926} & \textbf{0.953}\\
\hline\hline
\end{tabular}
\end{table}

\begin{table*}[t]
\caption{Open Set Recognition Results on CIFAR-10 with Various “Unknown” Datasets}
\label{table_2}
\centering
\begin{tabular}{l|c|c|c|c|c}
\hline\hline
Method & ImageNet-crop & ImageNet-resize & LSUN-crop & LSUN-resize & Avg. \\
\hline									
Softmax & 0.639 & 0.653 & 0.642 & 0.647 & 0.645\\
OpenMax \cite{Bendale2016} & 0.660 & 0.684 & 0.657 & 0.668 & 0.667\\
LadderNet+Softmax \cite{Yoshihashi2019} & 0.640 & 0.646 & 0.644 & 0.647 & 0.644\\
LadderNet+OpenMax \cite{Yoshihashi2019} & 0.653 & 0.670 & 0.652 & 0.659 & 0.659\\
DHRNet+Softmax \cite{Yoshihashi2019} & 0.645 & 0.649 & 0.650 & 0.649 & 0.648\\
DHRNet+OpenMax \cite{Yoshihashi2019} & 0.655 & 0.675 & 0.656 & 0.664 & 0.663\\
CROSR \cite{Yoshihashi2019} & 0.721 & 0.735 & 0.720 & 0.749 & 0.731\\
DOC \cite{Shu2017} & 0.760 & 0.753 & 0.748 & 0.764 & 0.756\\
MLOSR \cite{Oza2019a} & 0.837 & 0.826 & 0.783 & 0.801 & 0.812\\
CGDL \cite{Sun2020} & \textbf{0.840} & \textbf{0.832} & 0.806 & 0.812 & 0.823\\
OVRN-CD (Ours) & 0.835 & 0.825 & \textbf{0.846} & \textbf{0.839} & \textbf{0.836}\\
\hline\hline
\end{tabular}
\end{table*}

For the CIFAR-10 dataset, we followed the protocol in \cite{Yoshihashi2019}, where samples from ImageNet and LSUN were used as unknowns. The ImageNet and LSUN datasets were resized or cropped to make the unknown samples the same size as the known samples. Each dataset contains 10,000 testing images, and the known to unknown ratio is set to 1:1 during testing. TABLE \ref{table_2} shows that the proposed method performed the best, on average, providing the highest score for two unknown datasets. The comparison results show that setting class-specific sophisticated decision boundaries is key for high-performance OSR.

\section{Conclusion} \label{Chap5}
Alleviating the overgeneralization inherent in the closed set DNN classifier is the key to a high-performance OSR system. Thus, in this paper, we proposed a collective decision method that combines decisions reached by different OVRNs. The proposed method solves the overgeneralization problem by learning class-specific discriminative features and tightening decision boundaries, as verified through ablation studies. In addition, extensive comparison experiments were conducted on multiple standard datasets. The experimental results demonstrated that the proposed method outperforms state-of-the-art methods in many OSR scenarios, despite its simplicity.

We expect that more sophisticated decision boundaries can be set by additionally creating diverse synthetic samples that take up open space and learning these samples as unknowns. This approach will be addressed in our future work.

%\appendices

%\section{}

%\section*{Acknowledgment}


\ifCLASSOPTIONcaptionsoff
  \newpage
\fi



%\bibliographystyle{IEEEtran}

\bibliographystyle{IEEEtran}
\bibliography{Transactions-Bibliography/IEEEabrv,Transactions-Bibliography/reference}\ %IEEEabrv instead of IEEEfull

\clearpage

%\begin{thebibliography}{1}
%\end{thebibliography}


%\end{IEEEbiography}


%\newpage



%\vfill


\end{document}